\documentclass[10pt,letterpaper]{article}
\usepackage{aaai21}
\usepackage{times}
\usepackage{helvet}
\usepackage{courier}
\usepackage{pdfpages}
\usepackage{tabularx}
\usepackage{xparse}
\usepackage{amsmath}
\usepackage{amsthm}
\usepackage{amssymb}
\usepackage{xspace}
\usepackage{relsize}
\usepackage{graphicx}
\usepackage{multirow}
\usepackage[hyphens]{url}  
\usepackage{comment}
\usepackage{math_commands}
\usepackage{enumitem}
\usepackage{booktabs} 
\usepackage{tikz}
\usetikzlibrary{decorations.pathreplacing}
\usetikzlibrary{backgrounds}
\usepackage[linesnumbered,ruled,vlined]{algorithm2e}

\usepackage{mathabx}
\usepackage{mathrsfs}
\usepackage{xfrac}

\usepackage{multirow}

\usepackage{microtype}
\usepackage{subfig}

\usepackage{xr}

\makeatletter
\let\@myref\ref

\newcommand{\refsec}[1]{Sec.\,\@myref{#1}}
\newcommand{\refseq}[1]{Sec.\,\@myref{#1}}
\newcommand{\refig}[1]{Fig.\,\@myref{#1}}
\newcommand{\refigs}[2]{Fig.\,\@myref{#1}-\@myref{#2}}
\newcommand{\reftbl}[1]{Table \@myref{#1}}
\newcommand{\refstep}[1]{Step \@myref{#1}}
\newcommand{\refalgo}[1]{Alg. \@myref{#1}}
\newcommand{\refchap}[1]{Chapter \@myref{#1}}
\newcommand{\reflst}[1]{List \@myref{#1}}
\newcommand{\refeq}[1]{\@myref{#1}}

\makeatother

\newcounter{list}[section]

\usepackage{xcolor}

\newcommand{\RR}{\mathbb{R}}
\newcommand{\xmax}{x_{\max}} 
\newcommand{\GP}{\mathcal{GP}}
\newcommand{\N}{\mathcal{N}}
\newcommand{\D}{\mathcal{D}}
\newcommand{\y}{\mathbf{y}}
\newcommand{\ksi}{K+\eta^2 I}
\newcommand{\X}{\mathcal{X}}
\newcommand{\G}{\mathcal{G}}
\newcommand{\inv}{^{\raisebox{.2ex}{$\scriptscriptstyle-1$}}}
\newcommand{\ber}{\mathrm{Bernoulli}}
\newcommand{\ndim}{D}
\newcommand{\nsamples}{S}
\newcommand{\ncycles}{C}

\newcommand{\gs}{R}

\newcommand{\nlvl}{L}

\newcommand{\graphoverlap}{$\mathsf{Graph\ Overlap}$}
\newcommand{\graphnooverlap}{$\mathsf{Graph\ No\text{-}Overlap}$}
\newcommand{\graphnooverlapl}[1]{{$\mathsf{Graph\ No\text{-}Overlap~(#1)}$}}

\newcommand{\tree}{$\mathsf{Tree}$}
\newcommand{\random}{$\mathsf{Random}$}
\newcommand{\oracle}{$\mathsf{Oracle}$}

\DeclareMathOperator*{\fscore}{F_1score}
\DeclareMathOperator*{\precision}{Precision}
\DeclareMathOperator*{\recall}{Recall}

\DeclareMathOperator*{\Edge}{Edges}

\newcommand{\braces}[1]{{\left\{#1\right\}}}

\newcommand{\lsota}{state-of-the-art\xspace}  

\newcommand{\defun}[1]{%
\makeatletter
\expandafter\def\csname the#1\endcsname{\text{\it #1}}
\expandafter\def\csname #1\endcsname ##1{\csname the#1\endcsname\left(##1\right)}%
\makeatother
}

\newcommand{\defsetop}[2]{%
\makeatletter
 \expandafter\def\csname #1\endcsname ##1##2##3{%
  \expandafter\def\csname #1arg\endcsname{##1}%
  \expandafter\def\csname #1set\endcsname{##2}%
  \expandafter\def\csname #1cond\endcsname{##3}%
  \braces{##1##2\mid #2 ##3}%
 }%
\makeatother%
}

\defun{precond}
\defun{condition}
\defun{params}
\defun{type}
\defun{proc}
\defun{cost}

\defun{push}
\defun{pref}

\defun{init}
\defun{goal}
\defun{objects}
\defun{task}

\defun{parent}
\defun{plateau}

\defsetop{filter}{}
\defsetop{map}{}

\def\_{\\[-0.3em]}

\makeatletter

\newcommand{\newheuristic}[2]{%
 \def#1{%
  \ifmmode%
  h^\text{#2}\xspace%
  \else%
  \text{#2}\xspace%
  \fi%
 }%
}

\newheuristic{\lmcut}{LMcut}
\newheuristic{\mands}{M\&S}
\newheuristic{\pdb}{PDB}
\newheuristic{\ff}{FF}
\newheuristic{\ce}{CEA}
\newheuristic{\cg}{CG}
\newheuristic{\ad}{add}
\newheuristic{\lc}{LC}

\newcommand{\newUnitCostHeuristic}[2]{%
 \def#1{%
  \ifmmode%
  \hat{h}^\text{#2}\xspace%
  \else%
  \text{#2}\xspace%
  \fi%
 }%
}

\newUnitCostHeuristic{\lmcuto}{LMcut}
\newUnitCostHeuristic{\mandso}{M\&S}
\newUnitCostHeuristic{\ffo}{FF}
\newUnitCostHeuristic{\ceo}{CEA}
\newUnitCostHeuristic{\cgo}{CG}
\newUnitCostHeuristic{\ado}{add}
\newUnitCostHeuristic{\gco}{GoalCount}
\newUnitCostHeuristic{\lco}{LC}

\makeatother

\def\ref{\todo{Do not use ``ref'' directly!}}

\author{
    Eric Han,\textsuperscript{\rm 1}
    Ishank Arora,\textsuperscript{\rm 2}
    Jonathan Scarlett\textsuperscript{\rm 1,3}\\
}
\affiliations {
    \textsuperscript{\rm 1}School of Computing, National University of Singapore\\
    \textsuperscript{\rm 2}Indian Institute of Technology (BHU) Varanasi\\
    \textsuperscript{\rm 3}Department of Mathematics \& Institute of Data Science, National University of Singapore\\
    eric{\textunderscore}han@nus.edu.sg, 
    ishank.arora.cse14@iitbhu.ac.in, 
    scarlett@comp.nus.edu.sg
}

\title{High-Dimensional Bayesian Optimization via Tree-Structured Additive Models}

\begin{document}
\maketitle
\begin{abstract}
    Bayesian Optimization (BO) has shown significant success in tackling expensive low-dimensional black-box optimization problems. 
    Many optimization problems of interest are high-dimensional, and scaling BO to such settings remains an important challenge.
    In this paper, we consider generalized additive models 
    in which low-dimensional functions with overlapping subsets of variables are composed to model a high-dimensional target function.
    Our goal is to lower the computational resources required and facilitate faster model learning by \emph{reducing the model complexity}
    while retaining the \emph{sample-efficiency} of existing methods.
    Specifically, we constrain the underlying dependency graphs to tree structures in order to facilitate 
    both the structure learning and optimization of the acquisition function.
    For the former, we propose a hybrid graph learning algorithm based on Gibbs sampling and mutation.  In addition, we propose a novel zooming-based algorithm that permits generalized additive models to be employed more efficiently in the case of continuous domains.
    We demonstrate and discuss the efficacy of our approach via a range of experiments on synthetic functions and real-world datasets.
\end{abstract}

\section{Introduction}

Bayesian Optimization (BO) is a widespread method for sequential global optimization \cite{snoek2012practical}, 
and is suited to scenarios in which the target function $f$ is unknown and expensive to evaluate. 
BO was traditionally used in model selection \cite{movckus1975bayesian} and hyperparameter tuning \cite{snoek2012practical,swersky2013multi}.  
Recently, BO has also found success in black-box adversarial attack \cite{Ru2020BayesOpt}, robotics \cite{jaquier2020bayesian}, finance \cite{gonzalvez2019financial}, 
pharmaceutical product development \cite{sanoapplication}, natural language processing \cite{yogatama2015bayesian}, and more. 
Two critical ingredients of BO include a model that captures prior beliefs about the objective function, and an acquisition function that can be optimized efficiently.

BO has been most successful in low dimensions (i.e. $10$ or less) \cite{wang2013bayesian,nayebi2019framework}, 
whereas many applications require optimization in higher-dimensional spaces;
this remains a critical problem in BO \cite{wang2016practical,rolland2018high,frazier2018tutorial}.

A key difficulty associated with high-dimensional BO is the curse of dimensionality \cite{spruyt2014curse}, namely,
exponentially many observations are needed to find the global optimum in the absence of structural assumptions. 
Accordingly, two significant opposing challenges include the incorporation of suitable structural assumptions, and computationally efficient acquisition function optimization.
\subsection{Related Work}
In the literature, there are at least two approaches to high-dimensional BO with differing assumptions: 
\begin{itemize}[noitemsep,topsep=0pt,parsep=0pt,partopsep=0pt]
\item Under \textit{low effective dimensionality}, only few dimensions significantly affect $f$. 
\cite{chen2012joint} performed joint variable selection and optimization using GP-UCB.
\cite{djolonga2013high} applied low-rank matrix recovery techniques to learn the underlying effective subspace, and \cite{ijcai2019-0596} proposed a related approach based on sliced inverse regression.
\cite{wang2013bayesian} proposed REMBO, tackling the problem through random embedding.
More recently, \cite{kirschner2019adaptive} proposed LineBO, decomposing the problem into a sequence of one-dimensional sub-problems.  The use of {\em non-linear} low-dimensional embeddings has also recently been proposed \cite{lu2018structured,moriconi2019high}.
\item Under \textit{additive structure}, small subsets of variables interact with each other.
Specifically, additive models assume that $f$ can be decomposed into sums of lower-dimensional functions.
\cite{kandasamy2015high} assumed that the variables constructing a particular lower-dimensional function are not present in the other decomposed functions 
(i.e., the variables of each function are pairwise disjoint), which we refer to as \graphnooverlap.
\cite{rolland2018high} generalized the additive model to allow for an arbitrary dependency graph, removing the restriction of pairwise disjointness, which we refer to as \graphoverlap.
Also considering overlapping groups, \cite{hoang2018decentralized} assumed that \(f\) can be decomposed into several sparse factor functions, allowing for distributed acquisition function approximation.
\cite{li2016high} generalized to a projected-additive assumption; the model proposed by \cite{kandasamy2015high} is a special case when there is no projection.
Ensemble BO \cite{wang2018batched} seeks to not only exploit additive structures, but also use an ensemble of GP models through a divide and conquer strategy. 
\cite{NEURIPS2018_4e5046fc} combined additive GPs with approximations based on Fourier features, with the notable advantage of also establishing rigorous regret bounds.
\end{itemize}
In addition to the methods described above, other approaches have been taken to tackle high dimensionality.
\cite{li2017high} proposed a dropout strategy to optimize on a smaller subset of variables for every iteration.
\cite{oh2018bock} proposed BOCK, which tackles high-dimensionality via a cylindrical transformation of the search space.
\cite{NEURIPS2019_6c990b7a} proposed an approach based on running several local search procedures in parallel, and giving more samples to the most promising ones.
Other methods use deep neural networks combined with BO, such as \cite{snoek2015scalable,cui2019deep}.

The assumptions of low effective dimension vs.~additive structure are complementary. 
The performance of the optimization is dependent on the structure of the high-dimensional function, and trade-offs exist between computation time and accuracy.
Methods that assume low effective dimensionality are often computationally faster than additive methods; for example, due to scalability concerns, \cite{NEURIPS2019_6c990b7a} omitted methods that attempt to learn an additive decomposition from their experiments.
To the best of our knowledge, none of the existing works have scaled \graphoverlap~past 20 to 30 dimension.

In this paper, our focus is on additive structures; in particular, we seek to build on \graphoverlap. 
\graphoverlap~maintains computational tractability by using a message passing algorithm to optimize the acquisition function efficiently.
However, the message passing algorithm runs exponentially in the size of the maximum clique of the triangulated dependency graph \cite{rolland2018high}, impeding its scalability.

We see in the above-outlined works \cite{rolland2018high,hoang2018decentralized,li2016high} that the trend in the study of additive models has been to increase the model expressiveness.
An important caveat to such approaches is that a suitable model may be {\em much harder to find} given limited samples. Since one of the main premises of BO is optimizing with few samples, we contend that {\em simpler} models should also be sought to facilitate model learning with fewer samples, as well as reduced computation.

\subsection{Contributions}

The main contributions of this paper are as follows:
\begin{enumerate}[noitemsep,topsep=0pt,parsep=0pt,partopsep=0pt]
\item
  We trade-off expressiveness for scalability and ease of learning by reducing the complexity of the additive model, 
constraining the dependency structure to tree structures. As the function class is simpler, it reduces overfitting of the GP kernel,
and we are also able to reap computational efficiencies in both acquisition function optimization and dependency structure learning. 
\item 
  We propose a zooming technique for extending the message passing algorithm of \cite{rolland2018high} to continuous domains, 
  thus benefiting additive methods in general, and in particular our tree-based approach.
\item 
  We propose a hybrid method to learn the additive tree structures, composed of the following two techniques:
    \begin{enumerate}[noitemsep,topsep=0pt,parsep=0pt,partopsep=0pt]
      \item a tree structure growing algorithm that efficiently discovers edges that do not form cycles via Gibbs sampling;
      \item an edge mutation algorithm that obtains a new generation of trees from the current tree efficiently.
    \end{enumerate}
\item 
  Although limiting to tree structures may seem potentially risky due to the reduced expressivity, 
  we show this approach to be highly effective in a wide range of experiments, 
  indicating a highly competitive trade-off between expressive power and ease of model learning.
\end{enumerate}
We briefly mention that the use of tree structures in BO appeared in prior works \cite{jenatton2017tree,ma2020additive}, 
but with a very different type of model and motivation. These works aim to handle structured domains instead of real-valued domains, and
in contrast with our work, the tree represents binary decisions with only the leaves corresponding to Gaussian Processes.

\section{Additive GP-UCB using Tree Structures} 
We consider the sequential global optimization problem, 
seeking $\xmax = \arg \max_{x \in {\mathcal X}} f(x)$ for a \(\ndim\)-dimensional black-box function \(f: {\mathcal X} \to \RR\),
where ${\mathcal X} = \bigtimes_{i=1}^{\ndim}\X_i$ with each \(\X_i\) being an interval in \(\RR\). 
At the \(t\)-th observation, the algorithm selects $x_t$ and observes a noisy observation \(y_t = f(x_t) + \epsilon_t\), with \(\epsilon_t \sim\mathcal{N}(0,\eta^2)\). 

\subsection{Additive Dependency Tree Structures}
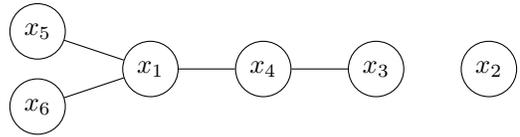
\begin{figure}
\centering
\begin{tikzpicture}[level distance=15mm,sibling distance=10mm,every node/.style={circle,draw}]
\node {$x_4$}
child[grow=left] {
    node{$x_1$}
child {node{$x_5$}} child {node{$x_6$}}
}
child[grow=right] {
node{$x_3$}
child {node{$x_2$} edge from parent[draw=none]}
};
\end{tikzpicture}
\caption{{Dependency tree structure, $h\left(x\right) = h^A(x_1, x_6) + h^B(x_1, x_5) + h^C(x_1, x_4) + h^D(x_3, x_4) + h^E(x_2)$.}\label{fig:dep-tree}}
\end{figure}
We use a Gaussian Process (GP) model to reason about the target function $f$.
Following \cite{rolland2018high}, we model \(f\) as a sum of several lower-dimensional components: 
\begin{equation}
    f\left(x\right) = \sum_{G\in \G} f^{G}\left(x^{G}\right),
    \label{eq:tree-add}
\end{equation}
where \(G \subseteq \{1, \dotsc, \ndim \}\)  denotes one subset of variables, and $\G$ represents the additive structure (see \refig{fig:dep-tree} for an example). 
The additive dependency structure is assumed to be tree-structured, possibly including forests. 
In contrast with \cite{rolland2018high}, in our setting the additive structure associated with any given graph is unique: 
Each lower-dimensional component \(f^G: \X^G \to \RR\) is either a 1 or 2-dimensional function defined on the variables in \(G\), 
where \(\X^G = \bigtimes_{v\in G} \X_v\).
Each edge represents a \(2\)-dimensional function, and each disconnected vertex represents a \(1\)-dimensional function.
\subsection{Prior and Posterior}
We model \(f\sim \GP\left(\mu, \kappa\right)\), with each \(f^{G}\) being an independent sample from a Gaussian Process \(\GP\left(\mu^G, \kappa^G \right)\), and 
\begin{equation}\begin{aligned}
\mu\left(x\right)&=\sum_{G \in \G}{\mu^G\left(x^G\right)}, \\
\kappa\left(x, x'\right)&=\sum_{G \in \G}{\kappa^G\left(x^G, {x'}^G\right)}.
\end{aligned}\label{eq:gp-prior}\end{equation}
We know from \cite{rolland2018high} that the posterior can be inferred via 
\(\left(f_*^G\mid\y\right) \sim \N\big(\mu_{t-1}^G, \left(\sigma_{t-1}^G\right)^2\big)\)
for each \(f^G_*\) at an arbitrary point \(x_*\) given \(\D_t=\{(x_i, y_i)\}^t_{i=1}\), 
where $\mathbf{y}=\left(y_1,\dotsc,y_t\right)$ correspond to $\mathbf{x}=\left(x_1,\dotsc,x_t\right)$, 
and the posterior mean and variance are given by
\begin{equation}\begin{aligned}
\mu_{t-1}^G&= \kappa^G\left(x_*^G, \mathbf{x}^G\right)\Delta\inv \y, \\
\left(\sigma_{t-1}^G\right)^2&= \kappa^G\left(x_*^G, x_*^G\right) \\&\quad-\kappa^G\left(x_*^G, \mathbf{x}^G\right)\Delta\inv\kappa^G\left(\mathbf{x}^G, x_*^G\right).
\end{aligned}\label{eq:gp-posterior}\end{equation}
Here we define the matrix
\(\Delta = \kappa\left(\mathbf x,\mathbf x\right)+\eta^2I_t \in \RR^{t\times t}\), \(\kappa\left(x_i,x_j\right)\) is the \((i,j)\)-th entry of \(\kappa\left(\mathbf x,\mathbf x\right)\), 
and $\kappa^G\left(\mathbf{x}^G, x_*^G\right)$ is of length $t$, with $i$-th entry $\kappa^G\left(x_i^G, x_*^G\right)$.

\section{Additive GP-UCB on Tree Structures}

\begin{algorithm}
\DontPrintSemicolon
Initialize $\D_0 \gets \left\{ \left(x_t, y_t\right) \right\}_{x_t \in X_\mathrm{init}}$\;
\For{$t=N_\mathrm{init} + 1,\dotsc,N_\mathrm{iter}$}{
    \If{$t \mod C=0$}{
        Learn $\G \gets$ {\sc Tree-Learning} (\refalgo{algo:tree-learning})\;
    }
    Update $\mu_{t}^G, \sigma_{t}^G : \forall {G\in\G}$ (\refeq{eq:gp-posterior})\;
    Optimize $x_t \gets \arg\max_{x \in\mathcal X} \phi_t\left(x\right)$ (\refalgo{algo:mp-tree}) \;
    Observe $y_t \gets f\left(x_t\right)+\epsilon$\;
    Augment $\D_t \gets \D_{t-1} \cup \left\{\left(x_t, y_t\right)\right\}$ \;
}
\Return $\arg\max_{\left(x,y\right)\in \D} y$\;
\caption{{\sc Tree-GP-UCB}}
\label{algo:add-gp-ucb-tree}
\end{algorithm}
In \refalgo{algo:add-gp-ucb-tree}, we present Tree-GP-UCB (\tree~for short).
Here, the total number of observations is \(N = N_\mathrm{init} + N_\mathrm{iter}\), 
where \(N_\mathrm{init}\) is the number of initial random samples \(X_\mathrm{init}\) drawn uniformly from \(\X\)
and \(N_\mathrm{iter}\) is the number of iterations.
For efficiency, \(\G\) and its hyperparameters are learned every \(\ncycles\) iterations, for some $\ncycles > 0$.

\subsection{Acquisition Function}
We focus on upper confidence bound (UCB) based algorithms \cite{auer2002using,srinivas2010gaussian}.
Specifically, following \cite{kandasamy2015high} and \cite{rolland2018high}, we let the global acquisition function \(\phi_t(x)\) be the sum of the individual acquisition functions with respect to the dependency structure \(\G\):
\begin{equation}\begin{aligned}
\phi_t\left(x\right) &= \sum_{G\in\G}{\phi^G_t\left(x^G\right)},\\
\phi^G_t\left(x^G\right)&=\mu_{t-1}^G\left(x^G\right)+\beta_t^{{1}/{2}}\sigma_{t-1}^G\left(x^G\right).\\
\end{aligned}\label{eq:acquisition-gp-tree}\end{equation}

\subsubsection{Maximization over Continuous Domains.}
The message passing approach proposed by \cite{rolland2018high} works on discrete domains.
A naive approach to handle continuous domains would be to discretize the continuous domain uniformly (i.e., a grid with equal spacing).
However, this may require large amounts of computation, especially when the discretization is performed using a small spacing. 
Here, we present a refined message passing algorithm specifically designed for continuous domains.

\begin{algorithm}
\DontPrintSemicolon
Initialize $\left(\mathbf a, \mathbf b\right)$ with the bounds of $\X$ \;
\For{$l=1,\dotsc,\nlvl$}{
 \For{$d=1,\dotsc,\ndim$}{
 Discretize $\X_d \gets [[a_d, b_d]]_\gs$ \tcp*{$\left|\X_d\right| = \gs$}
 }
 $\X \gets \bigtimes_{d=1}^{\ndim}\X_d$\;
 $\left(x,y\right) \gets \text{\sc Msg-Passing-Discrete} \left(\X\right)$\;
 Select $\left(\mathbf a, \mathbf b\right) \gets \text{\sc Zoom-Strategy}\left(x\right)$\;
}
\Return $\left(x,y\right)$\;
\caption{{\sc Msg-Passing-Continuous}}
\label{algo:mp-tree}
\end{algorithm}

The optimization of the acquisition function over continuous domains is presented in \refalgo{algo:mp-tree};
it starts with the full continuous domain
\(\X = \bigtimes_{d=1}^{\ndim}\X_d\), where \(\X_d \in \left [a_d, b_d\right] \subseteq \R \). 
Firstly, we discretize each variable's domain to a finite subset, and let 
$\gs$ denote the size of the subset.
Thereafter, we use a simplified version of the message passing algorithm {\sc Msg-Passing-Discrete}
of \cite{rolland2018high}---\refalgo{algo:mp-tree-discrete} in the appendix---to perform optimization over the discretized domain. 
As the dependency graph is a tree, the complexity of message passing is quadratic in $\gs$. 
The bounds \(\left(\mathbf a, \mathbf b\right)\) for the next level are picked by \text{\sc Zoom-Strategy} (see below) given the selected point. 
We perform the steps iteratively for some number \(\nlvl\) of levels.
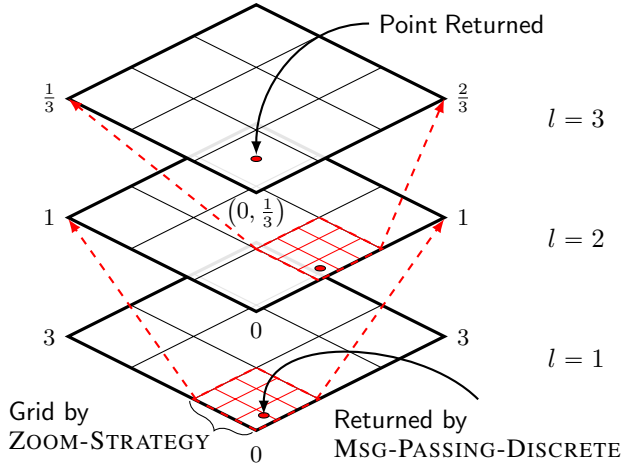
\begin{figure}
    \centering
    \begin{tikzpicture}[scale=.5]
        
        \begin{scope}[
            yshift=-180,every node/.append style={
            yslant=0.5,xslant=-1},yslant=0.5,xslant=-1]
            \fill[white,fill opacity=.9] (0,0) rectangle (5,5);
            \draw[step=16.67mm, black] (0,0) grid (5,5);
            \draw[step=5.56mm,red] (0,0) grid (1.67,1.67);
            \draw[black,very thick] (0,0) rectangle (5,5);
            \draw[red,thick,dashed] (0,0) rectangle (1.67,1.67);
            
            \draw [fill=red](0.5,0.3) circle (.1) ;
            \draw [decorate,decoration={brace,amplitude=6pt}](-0.1,0) -- (-0.1,1.67);    
        \end{scope} 
        \begin{scope}[
            yshift=-90,every node/.append style={
            yslant=0.5,xslant=-1},yslant=0.5,xslant=-1
            ]
            \fill[white,fill opacity=.9] (0,0) rectangle (5,5);
            \draw[step=16.67mm, black] (0,0) grid (5,5);
            \draw[step=5.56mm,red] (1.66,0) grid (3.34,1.67);
            \draw[black,very thick] (0,0) rectangle (5,5);
            \draw[red,thick,dashed] (1.66,0) rectangle (3.34,1.67);
            \draw [fill=red](2,0.3) circle (.1) ;
        \end{scope}
    
        \begin{scope}[
            yshift=0,every node/.append style={
                yslant=0.5,xslant=-1},yslant=0.5,xslant=-1
                         ]
            \fill[white,fill opacity=.9] (0,0) rectangle (5,5);
            \draw[step=16.67mm, black] (0,0) grid (5,5);
            \draw[black,very thick] (0,0) rectangle (5,5);
            \draw [fill=red](0.9,0.9) circle (.1) ;
        \end{scope}
        
        \draw[-latex,thick,dashed,red] (3.3,-1.5) to (5,2.5);
        \draw[-latex,thick,dashed,red] (0,-1.5) to (-5,2.5);
    
        \draw[-latex,thick,dashed,red] (1.6,-5.5) to (5,-0.7);
        \draw[-latex,thick,dashed,red] (-1.6,-5.5) to (-5,-0.7);
    
        \draw (7.5,-4.4) node[right] {$l=1$};
        \draw (7.5,-1.2) node[right] {$l=2$};
        \draw (7.5,2) node[right]{$l=3$};
    
        \draw[-latex,thick,black](5.9,-5.5) node[below,text width=3.85cm]{$\mathsf{Returned\ by\ }$\\ $\text{\sc Msg-Passing-Discrete}$}
         to[out=135,in=90] (0.2,-5.9);

        \draw (-1,-6.2) node[left,text width=2.67cm]{$\mathsf{Grid\ by\ }$\\ $\text{\sc Zoom-Strategy}$};

        \draw[-latex,thick,black](3,4.5) node[right]{$\mathsf{Point\ Returned}$}
            to[out=180,in=90] (0,1);
    
        \fill[black,font=\footnotesize]
            (-5.5,-4.3) node [above] {$3$}
            (0,-6.5) node [below] {$0$}
            (5.5,-4.3) node [above] {$3$};
        \fill[black,font=\footnotesize]
            (-5.5,-1.1) node [above] {$1$}
            (0,-3.2) node [below] {$0$}
            (5.5,-1.1) node [above] {$1$};
        \fill[black,font=\footnotesize]
            (-5.5,1.9) node [above] {$\frac{1}{3}$}
            (0,0.1) node [below] {$\left(0, \frac{1}{3}\right)$}
            (5.5,1.9) node [above] {$\frac{2}{3}$};
    \end{tikzpicture}
{\caption{Example with two variables, grid size \(\gs=3\), and domain \([0,3]\).
Firstly, we partition each axis evenly into \(3\) partitions. 
Next, we draw a uniformly random point from each partition. 
The points from each axis form a discretized domain, and we run \text{\sc Msg-Passing-Discrete} on this discrete domain. 
Finally, we zoom into the square, representative of the selected point.
In this manner, we recursively sub-divide the grid for all \(\nlvl=3\) levels.}\label{fig:mp-cont-example}}
\end{figure}
\paragraph{Zoom Strategy.}
Different strategies can be employed in choosing the bounds and their representative points for the next level. 
We adopt a simple randomized strategy exemplified in \refig{fig:mp-cont-example}: 
At each level, we partition each current interval \([a_i, b_i]\) uniformly onto a grid of size \(\gs\), 
and choose a uniformly random point within that interval as its representative. 
We refer to this discretization of the domain as $[[a_i, b_i]]_\gs$.
We use \text{\sc Msg-Passing-Discrete} restricted to these representatives, and for the one chosen, we recursively zoom into the corresponding sub-domain.
Henceforth, we use {\sc Msg-Passing-Continuous} with this zoom strategy.

\subsection{Additive Components}
The choice of an appropriate kernel and the learning of its parameters are critical to the success of BO. 
In high-dimensional additive BO, the problem compounds, 
as we need to learn the dependency structure along with kernel parameters for every kernel in the additive model.

As mentioned previously, an additive decomposition \(\G\) corresponds to a dependency graph;
the additive function \(f\left(x\right) = \sum_{G\in \G} f^{G}\left(x^{G}\right)\) is the sum over its additive components in $\G$.  It will be convenient to work with the equivalent representation of an adjacency matrix \(Z\in \{0,1\}^{\ndim\times \ndim}\), where \(Z_{ij}=1\) if variables \(i\) and \(j\) are connected on the (tree-structured) graph. 
Assuming that each function's kernel \(\kappa^G\) is parameterized by some kernel parameters \(\theta^G\) (e.g., lengthscale etc.), 
the overall collection of parameters is \(\Theta_{\G}=\left\{\theta^G\right\}_{G\in\G}\) given a decomposition \(\G\). 
We note that learning the kernel parameters along with the decomposition \(\G\) is difficult,
as the search space is large and we may encounter problems with overfitting. 
We tackle this problem by defining a fixed set of dimensional kernel parameters $\Theta$ that are independent of the decomposition 
and defining the kernel parameters over them; see \refsec{sec:kernel} for details.

\paragraph{Maximum likelihood.}
For model learning, we make use of the maximum log-likelihood score, given by
\begin{equation}
\begin{aligned}\rho\left(Z,\theta\right) = 
 &-\frac{1}{2}\y^T \left(\ksi\right)^{-1} \y \\
 &- \frac 1 2 \log |\ksi| - \frac{n} 2 \log 2\pi,\label{eq:lml}
\end{aligned}
\end{equation}
where \(K\in\RR^{n\times n}\) is the kernel matrix of the observed
data points \(n\), assuming a dependency graph \(\G\) with an equivalent
adjacency matrix \(Z\) and parameters \(\theta\).

\subsubsection{Dependency Structure Learning.}

Following \cite{wang2017batched,rolland2018high}, we adopt a Bayesian approach to structure learning, 
on which we place a prior distribution on $Z$ and seek to sample from the posterior distribution. 
We use Gibbs sampling to sample approximately, 
avoiding the difficult task of sampling directly from the high-dimensional distribution over tree structures.

Specifically, we use such sampling to update the presence/absence of edges from variable \(i\) to \(j\), but to maintain the tree structure, we discard edges that would create a cycle.
We assume a prior with Bernoulli random variables with parameter \(\gamma\), \(Z_{ij}\sim \ber\left(\gamma\right)\). 
We can use this model to formulate the posterior for \(Z_{ij}\); letting $\mathcal{D}$ denote the data collected, and letting $Z_{-(ij)}$ be the adjacency variables excluding $(i,j)$, we have the following \cite{rolland2018high}:
\begin{equation}
    P\left(Z_{ij}=1\mid Z_{-(ij)},\theta,\mathcal{D}; \gamma\right) \propto \gamma \, e^{\rho\left(Z_{ij}=1\cup Z_{-(ij)},\theta\right)}.
\end{equation}
For each \(Z_{ij}\), we compare the log of the posterior for two cases:
\(\log\left(\gamma\right)+\rho\left(Z_{ij}=1\cup Z_{-ij},\theta\right)\)
vs.~\(\log\left(1-\gamma\right)+\rho\left(Z_{ij}=0 \cup Z_{-ij},\theta\right)\).
The parameter \(\gamma\) can be set to \(1/2\) if there is no prior information about \(Z\). 
We use the log-likelihood in two ways, combining them to learn the structure in \refalgo{algo:tree-learning}. 
First, we use Gibbs Sampling to build a connected tree from an empty graph iteratively.
Once the dependency graph is a connected tree, we apply mutation in subsequent iterations. 
Thus, we grow the empty graph into a tree and then seek improvements via mutation.

\begin{algorithm}
\DontPrintSemicolon
$\mathcal{Z} \gets \{Z^\mathrm{current}\}$ \;
$Z^{(k)} \gets Z^\mathrm{current}$\;
\While{$k<\nsamples $}{
 \If{$\text{\sc Number-Of-Edges}\left(Z^{(k)}\right) < \ndim -1$} {
    Update $(\mathcal{Z},k)$ via {\sc Gibbs-Sampling} (\refalgo{algo:gibbs-tree})\;
 } \Else{
    Update $(\mathcal{Z},k)$ via {\sc Mutation} (\refalgo{algo:gibbs-mutation})\;
 }
}
\Return $Z\in \mathcal{Z}$ with the highest likelihood score\;
\caption{{\sc Tree-Learning}}
\label{algo:tree-learning}
\end{algorithm}
\paragraph{Adding Edges.}
\refalgo{algo:gibbs-tree} samples from the marginal posteriors, while only adding edges that maintain that \(Z\) is still a tree. 
The Union-Find (UF) data structure tracks a set of disjoint sets, providing the operations \emph{union} and \emph{find}.
Both operations can be performed in (amortized) time $O\left(\alpha\left(\ndim\right)\right)$ when implemented using weights with path compression \cite{cormen2009introduction,sedgewick2011algorithms},
where $\alpha\left(\ndim\right)$ is the inverse Ackermann function. 
In short, both operations can be performed in nearly constant time (amortized).
In our algorithm, we use UF to track the connected components of $\G$, represented by disjoint subsets of variables.
We use the find operation to check for cycles. 
After adding the edge, we update UF by performing the union operation.

\begin{algorithm}
\DontPrintSemicolon
$\text{Initalize UF data structure}$\;
\For{$j=1,\dotsc,\ndim$}{
 \For{$i=1,\dotsc,j-1$}{
 $Z^{(k+1)} \gets Z^{(k)}$\;
 \If{$\text{cycle not formed by } Z_{ij}^{{(k+1)}}=1$} {
 Sample $Z_{ij}^{\text {(new)}}$ from posterior \;
 $Z^{(k+1)} \gets Z_{ij}^{\text {(new)}}$\;
 $\text{Update UF via union operation}$\;
 Add $\mathcal{Z} \gets \mathcal{Z} \cup\left\{Z^{(k+1)}\right\}$\; 
 }
 $k \gets k + 1$ \;
 }
}
\caption{{\sc Gibbs-Sampling} at $k$-th iteration}
\label{algo:gibbs-tree}
\end{algorithm}
\paragraph{Mutation.}
\refalgo{algo:gibbs-mutation} describes the mutation operation that we perform when the dependency graph \(\G\) is a connected tree. 
We borrow the idea of mutation from genetic algorithms; the mutation operation can maintain tree structure diversity from one generation to another. 
The purpose of the mutation operation is to preserve and introduce diversity, 
wherein genetic algorithms, a mutation helps to avoid getting stuck in local maxima by making minor changes to the previous generation. 
In our context, the population is a new generation of trees in each iteration, and the fitness function is the log-likelihood.
Using mutation, we can simultaneously avoid local maxima and efficiently maintain a tree structure.

We note that one could simply use the Gibbs sampling approach or the mutation approach separately rather than using the former followed by the latter, but we found this combined approach to be effective experimentally.

\begin{algorithm}
\DontPrintSemicolon
$Z^{(k+1)} \gets Z^{(k)}$\;
$i,j \gets$ Sample random edge for which $Z_{ij}^{{(k+1)}}=1$\;
Remove edge: $Z_{ij}^{{(k+1)}}=0$\;
$i',j' \gets$ Sample nodes from the disconnected sub-trees\;
Sample $Z_{i'j'}^{\text {(new)}}$ using posterior\;
$Z^{(k+1)} \gets Z_{i'j'}^{\text {(new)}}$\;
Augment the dataset: $\mathcal{Z} \gets \mathcal{Z} \cup\left\{Z^{(k+1)}\right\}$\;
$k \gets k + 1$ \;
\caption{{\sc Mutation} at $k$-th iteration}
\label{algo:gibbs-mutation}
\end{algorithm}

\section{Experimental Results} 

For each of our experiments,\footnote{The code is available at \url{https://github.com/eric-vader/HD-BO-Additive-Models}.} we compare our method, 
\tree,~to several \lsota~black-box global optimization methods, particularly BO methods including \graphnooverlap~\cite{kandasamy2015high}, 
\graphoverlap~\cite{rolland2018high}, LineBO \cite{kirschner2019adaptive}, and REMBO \cite{wang2013bayesian}. 

To avoid clutter, we avoid including every algorithm and baseline in our charts. 
Instead, we make an effort to compare against the \emph{best} algorithm for the function at hand,  leveraging on prior works' experimental results to complete our discussion.
For example, in cases where LineBO was already shown to outperform standard GPs and REMBO in \cite{kirschner2019adaptive}, we omit these worse-performing methods.

We run all additive methods using the zooming-based message passing algorithm, analogous to \refalgo{algo:mp-tree}.
In addition, we compare to \random, which evaluates points at random.
Where possible, we also compare our results to \oracle, which has access to the true dependency graph along with the true kernel parameters. 
The functions and data sets considered are summarized in \reftbl{tab:dataset-summary} in the appendix.

\subsection{Setup}
\label{sec:setup}
Whenever possible, we used identical parameters across all competing algorithms and functions.
However, we note that most algorithms have unique hyperparameters. 
We set those hyperparameters to reasonable values, discussed in the appendix.
The competing algorithms and their unique hyperparameters are given in \reftbl{tab:sota-summary} in the appendix.
We ran each algorithm \(25\) times for every function with varied conditions.\footnote{Conditions include initial points, instances of the objective function, and random seeds used by the algorithm.} 
We ran all experiments with $N_\mathrm{init}=10$ initial points and $N_\mathrm{iter}=1000$ total points.
The same conditions are used across all algorithms to ensure a fair comparison. 

\paragraph{Kernel.\label{sec:kernel}} We adopt the widely-used Radial Basis Function (RBF) kernel, 
more specifically using a variant known as the RBF-ARD kernel \cite{murphy2012machine},
which consists of a dimensional lengthscale \(\ell_i\) for every dimension \(i\). 
In addition, we decompose the scale parameter $\sigma^G=\sqrt{\sum_{i\in G}{\sigma_i}^2}$ into its dimensional components \(\sigma_i\), 
so that we can learn the parameters tractably. 
Each low-dimensional kernel corresponds to a low-dimensional function with set of variables \(G\):
\begin{equation}\kappa^G_\mathrm{RBF}\left(x,x'\right) = \sigma^G \exp\left(-\frac{1}{2}\sum_{i\in G}\frac{(x_i-x_i')^2}{\ell_i^2}\right).
\label{eq:rbf-ls}\end{equation}
In this manner, the kernel parameters $\Theta_\G$ are defined over the dimensional kernel parameters 
$\Theta = \left\{\left(\ell_i, \sigma_i\right)\right\}_{i=1}^{\ndim}$. 
We adopt the established gradient-based approach to learning $\Theta$; see \refsec{sec:kernel-param} in the appendix.
We initialize the dimensional lengthscale and scale parameters as $\sigma_i = 0.5$, and $l_i = 0.1$ for all $i$.
We set $\eta=0.1$ in (\refeq{eq:gp-posterior}) to account for noisy observations.

\paragraph{Additive Models.} 
All additive models start with an empty graph of the appropriate size for the given function. 
Concerning the learning of the dependency structure, we assume no prior knowledge (\(\gamma=0.5\)). 
We sample the structure for \(\nsamples = 250\) times every $\ncycles=15$ iterations. 
After learning the structure, we choose the best kernel parameters using the gradient approach mentioned above.
We set the trade-off parameter in UCB to be \(\beta\left(t\right) = 0.5\log{\left(2t\right)}\), 
as suggested in \cite{rolland2018high}. 
For \textit{discrete} experiments, we discretize each dimension to \(50\) levels, with the
maximum number of individual acquisition function evaluations capped at \(1000\). 
For \textit{continuous} experiments, we let each level's grid size be \(\gs=4\) and the number
of levels be \(\nlvl=4\) (see \refig{fig:mp-cont-example}) with no maximum evaluation limits.

\subsection{Metrics}
Following \cite{wang2013bayesian}, we plot the mean and $\sfrac{1}{4}$ standard deviation confidence intervals 
of the metrics over all $25$ runs of the algorithm. 
For convenience, each plot's legend is ordered according to the curves' final $y$-value.

\subsubsection{Graph Learning Performance.}

We measure how close the estimate $G$ is from its target graph $G_\text{opt}$ by calculating  
\begin{equation}
    \fscore{\left(G\right)} = 2 \frac{\precision{\left(G\right)} \times \recall{\left(G\right)}}{\precision{\left(G\right)} + \recall{\left(G\right)}},
\end{equation}
where $\precision{\left(G\right)} = \frac{\left|\Edge{\left(G\right)} \cap \Edge{\left(G_\text{opt}\right)}\right|}{\left|\Edge{\left(G\right)}\right|}$ and $\recall{\left(G\right)} = \frac{\left|\Edge{\left(G\right)} \cap \Edge{\left(G_\text{opt}\right)}\right|}{\left|\Edge{\left(G_\text{opt}\right)}\right|}$, with 
$\Edge{\left(G\right)}$ denoting the set of edges in graph $G$. A larger $\fscore$ indicates better graph learning performance.

\subsubsection{Optimization Performance.}
In accordance with the ultimate goal of BO, we compute the \emph{best regret} to measure closeness to the best value $f_\mathrm{max}$ at iteration \(i\):
\begin{equation}
R_t = f_\mathrm{max} - f_i^*
\label{eq:avg-cum-best-regret},
\end{equation}
where \(f_i^*\) denotes the best \(f\left(x\right)\) value sampled up to iteration \(i\). 
For functions where \(f_\mathrm{max}\) is unknown, we instead consider \(f_i^*\), i.e., the \emph{best value} found.
\begin{figure*}[!ht]
    \centering
    \subfloat[Star-25 (Discrete) Performance]{\includegraphics[width=0.32\textwidth]{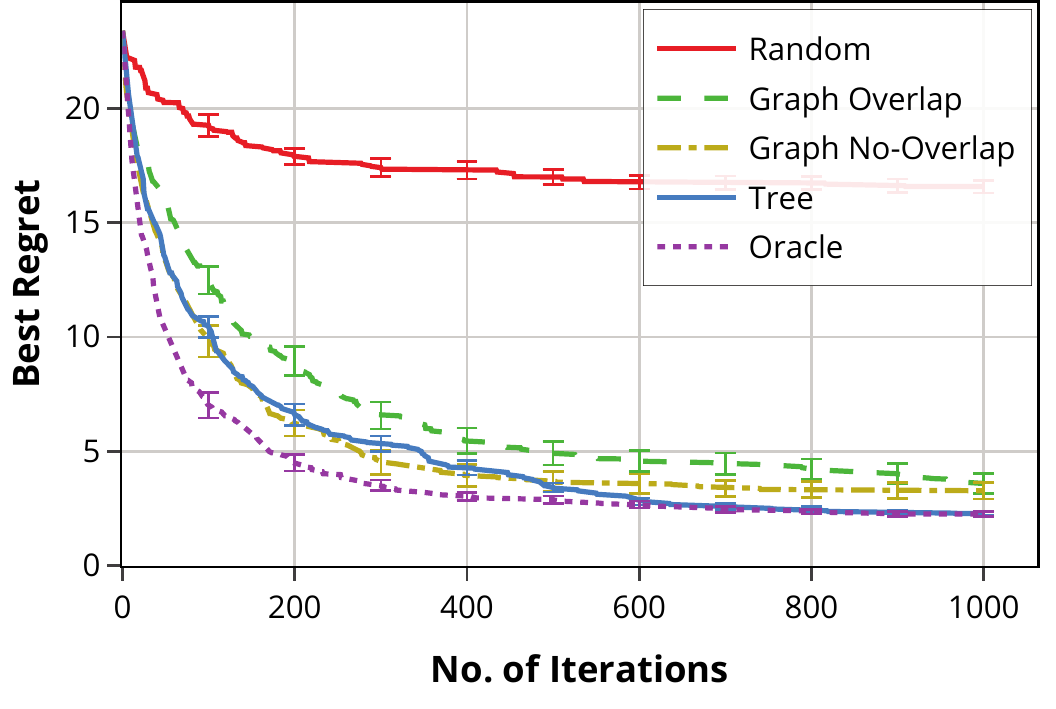} \label{fig:StarGraph-25-synd-performance}}
    \subfloat[Grid-3$\times$3 (Continuous)  Performance]{\includegraphics[width=0.32\textwidth]{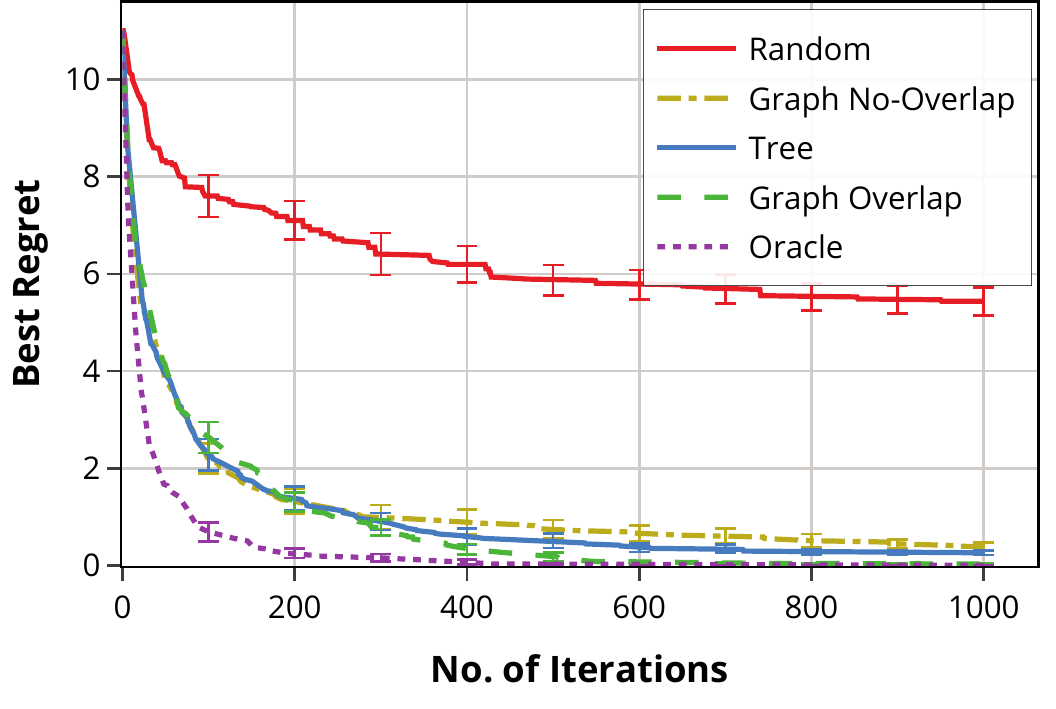} \label{fig:GridGraph-9-sync-performance}}
    \subfloat[Grid-3$\times$3 (Continuous)  $\fscore$]{\includegraphics[width=0.32\textwidth]{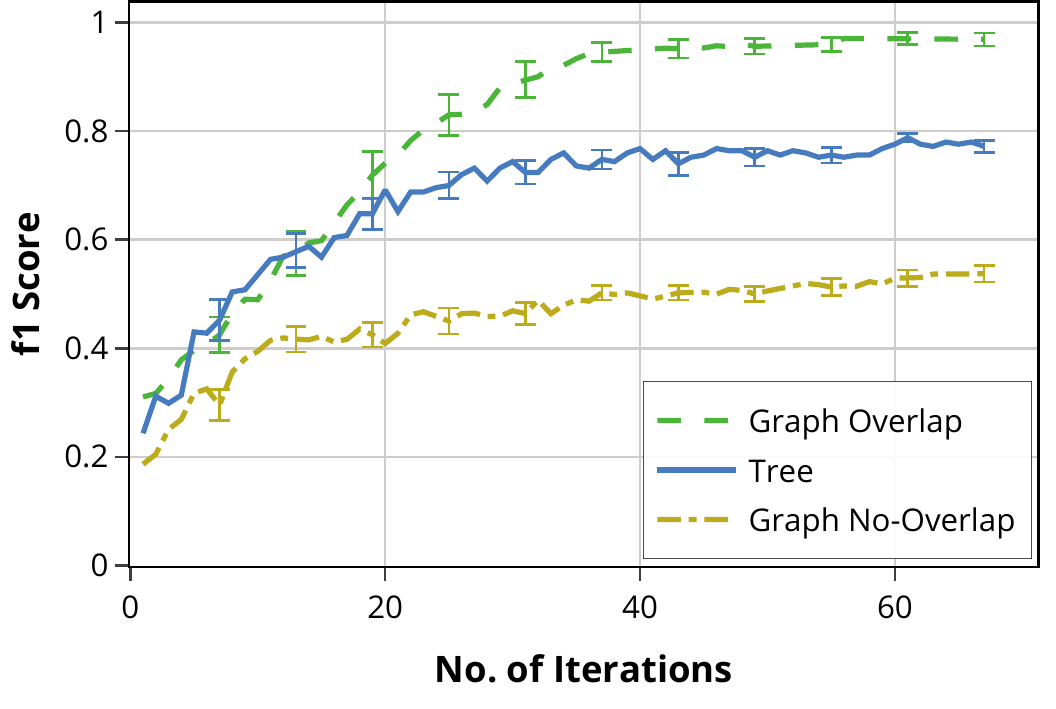} \label{fig:GridGraph-9-sync-cg}}\\
    \subfloat[Ancestry-132 (Continuous)  Performance]{\includegraphics[width=0.32\textwidth]{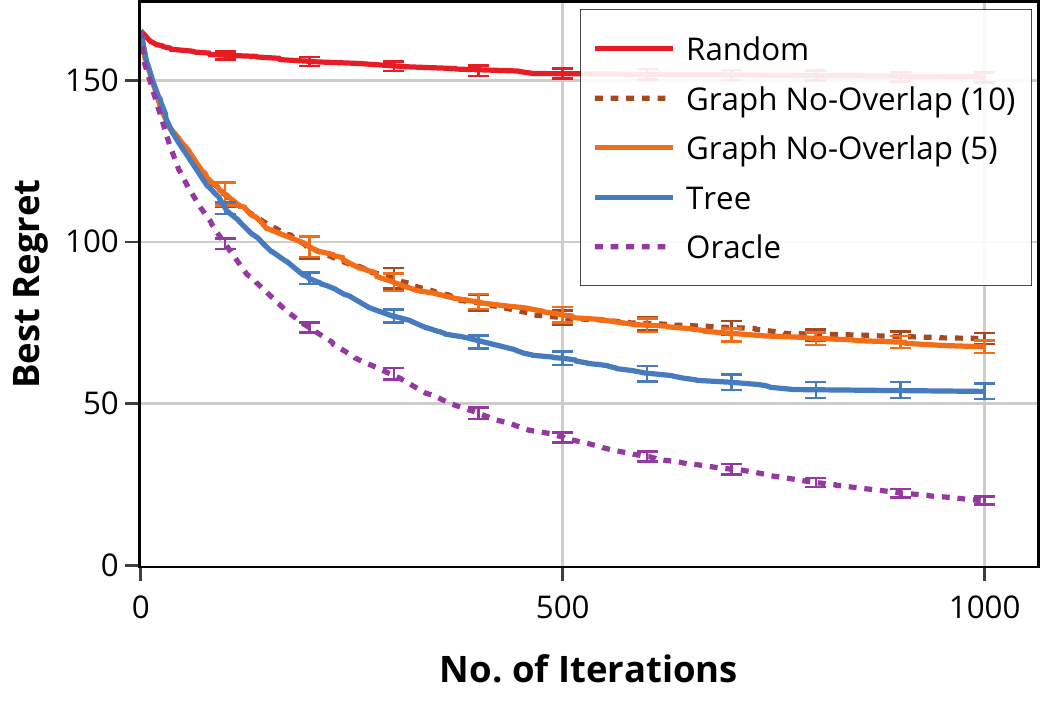} \label{fig:AncestryGraph-132-sync_tree-performance}}
    \subfloat[Ancestry-132 (Continuous)  Cost]{\includegraphics[width=0.32\textwidth]{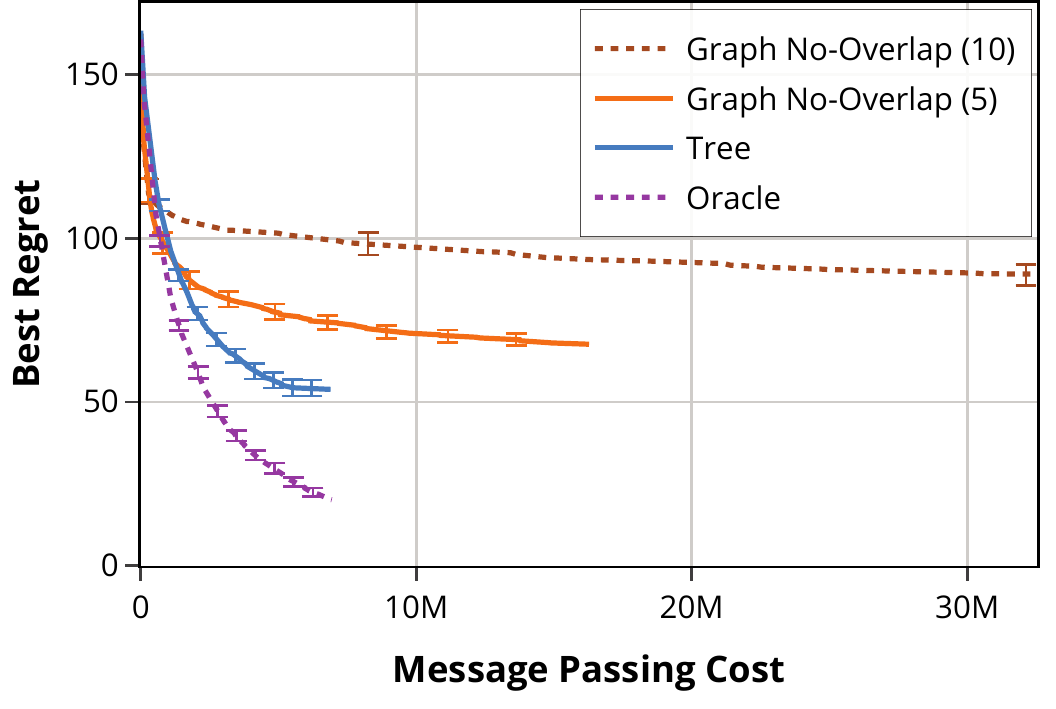} \label{fig:AncestryGraph-132-sync_tree-cost}}
    \subfloat[Scalability of \tree~over dimensions]{\includegraphics[width=0.32\textwidth]{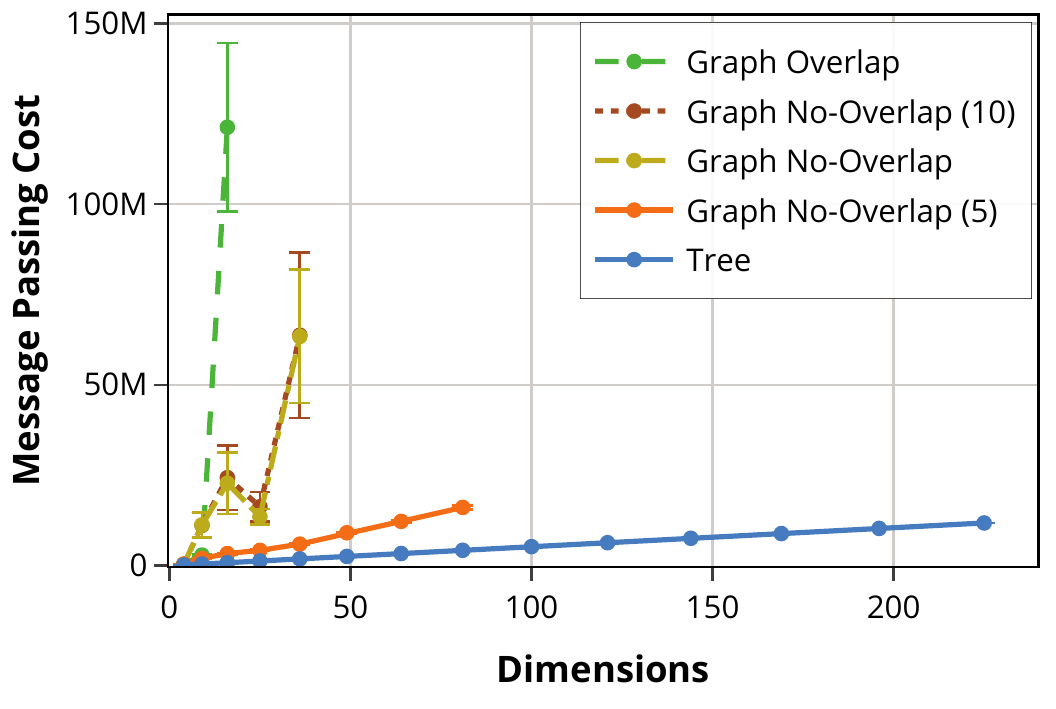}\label{fig:scalability}}
    \caption{Summarized comparison of various additive methods across various functions.}
\end{figure*}

\begin{figure}
    \centering
    \subfloat[Star-25]{
    \begin{tikzpicture}
        \tikzstyle{every node}=[circle, draw, fill=black!50,
        inner sep=0pt, minimum width=4pt]
        \node at (360:0mm) (center) {};
        \foreach \n in {1,...,24}{
            \node at ({\n*360/24}:1.02cm) (n\n) {};
            \draw (center)--(n\n);
        }
    \end{tikzpicture}}
    \qquad
    \subfloat[Partition-12]{
    \begin{tikzpicture}
        \tikzstyle{every node}=[circle, draw, fill=black!50,
        inner sep=0pt, minimum width=4pt]

        \def\len{0.8}
        \def\cos{\len*cos(60)}
        \def\cosv{{\cos}}
        \def\sin{\len*sin(60)}
        \def\sinv{{\sin}}
        \node at (0,0) (A) {};
        \node at (\len,0) (B) {};
        \node at (\cosv,\sinv) (C) {};
        \draw (A)--(B);
        \draw (B)--(C);
        \draw (A)--(C);

        \node at ({2-\len},0) (D) {};
        \node at (2,0) (E) {};
        \node at ({2-\len+\cos},\sinv) (F) {};
        \draw (D)--(E);
        \draw (E)--(F);
        \draw (D)--(F);

        \node at ({2-\len},{2-\sin}) (G) {};
        \node at (2,{2-\sin}) (H) {};
        \node at ({2-\len+\cos},2) (I) {};
        \draw (G)--(H);
        \draw (H)--(I);
        \draw (G)--(I);

        \node at (0,{2-\sin}) (J) {};
        \node at (\len,{2-\sin}) (K) {};
        \node at (\cosv,2) (L) {};
        \draw (J)--(K);
        \draw (K)--(L);
        \draw (J)--(L);

    \end{tikzpicture}}
    \qquad
    \subfloat[Grid-3$\times$3]{
    \begin{tikzpicture}
        \tikzstyle{every node}=[circle, draw, fill=black!50,
        inner sep=0pt, minimum width=4pt]
        \def\maxX{2}
        \foreach \x [count=\n] in {0,...,\maxX}{
            \foreach \y in {0,...,2}{
                \node at (\x,\y) (N{\x,\y}) {};    
            }
        }
        \foreach \x [count=\n] in {0,...,\maxX}{
            \draw (N{\x,0}) -- (N{\x,1});
            \draw (N{\x,1}) -- (N{\x,2});
        }
        \foreach \y in {0,...,2}{
            \draw (N{0,\y}) -- (N{1,\y});
            \draw (N{1,\y}) -- (N{2,\y});
        }
    \end{tikzpicture}}
    {\caption{Synthetic Dependency Graphs Structures.}\label{fig:synthetic}}
\end{figure}
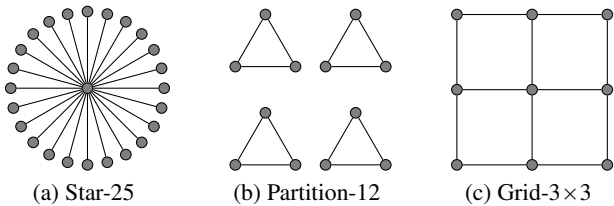
\begin{figure*}
    \centering
    \subfloat[Hartmann6+14Aux Performance]{\includegraphics[width=0.32\textwidth]{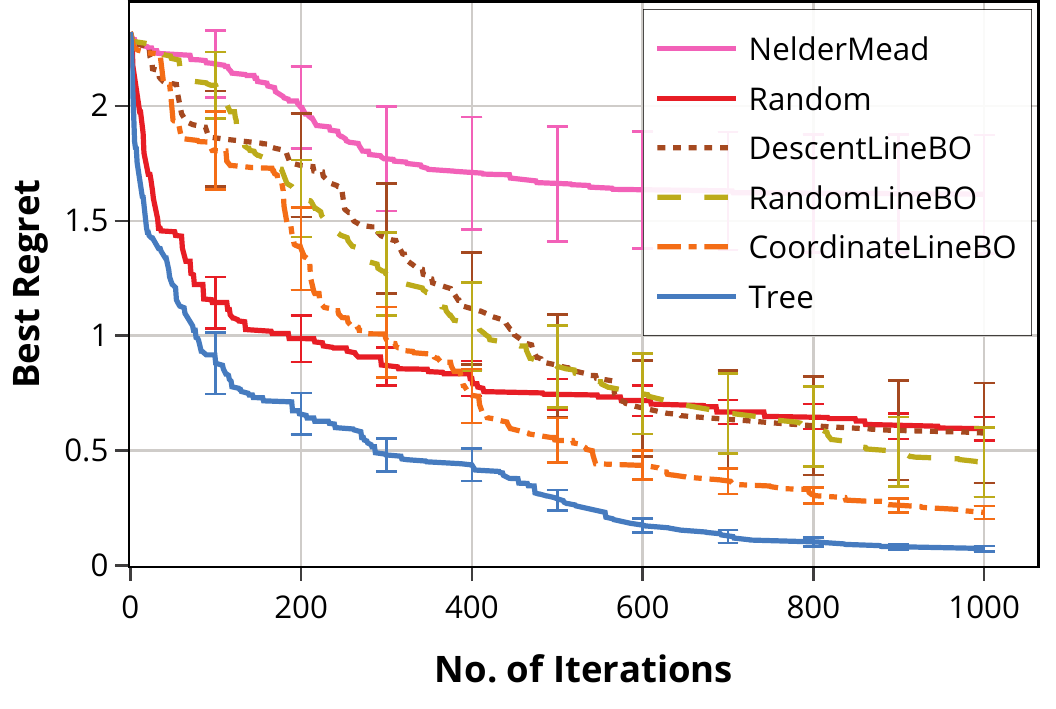} \label{fig:Hartmann6Aug-hpoca-performance}}
    \subfloat[Stybtang250 Performance]{\includegraphics[width=0.32\textwidth]{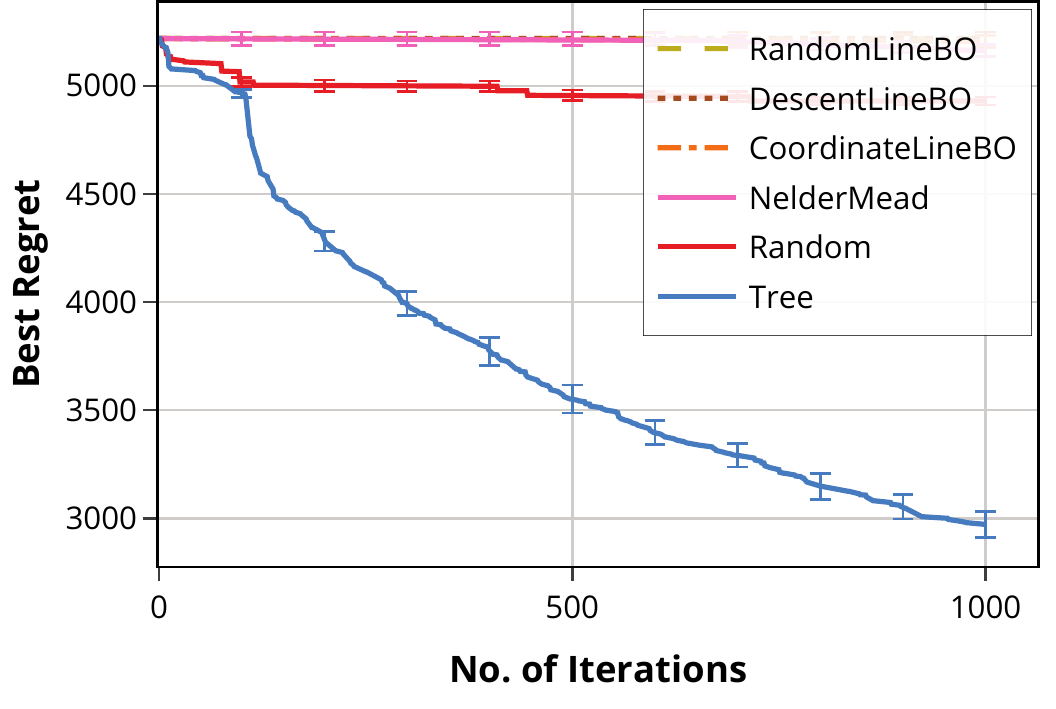} \label{fig:Stybtang-stybtang-performance}}
    \subfloat[Lpsolve-misc05inf Performance]{\includegraphics[width=0.32\textwidth]{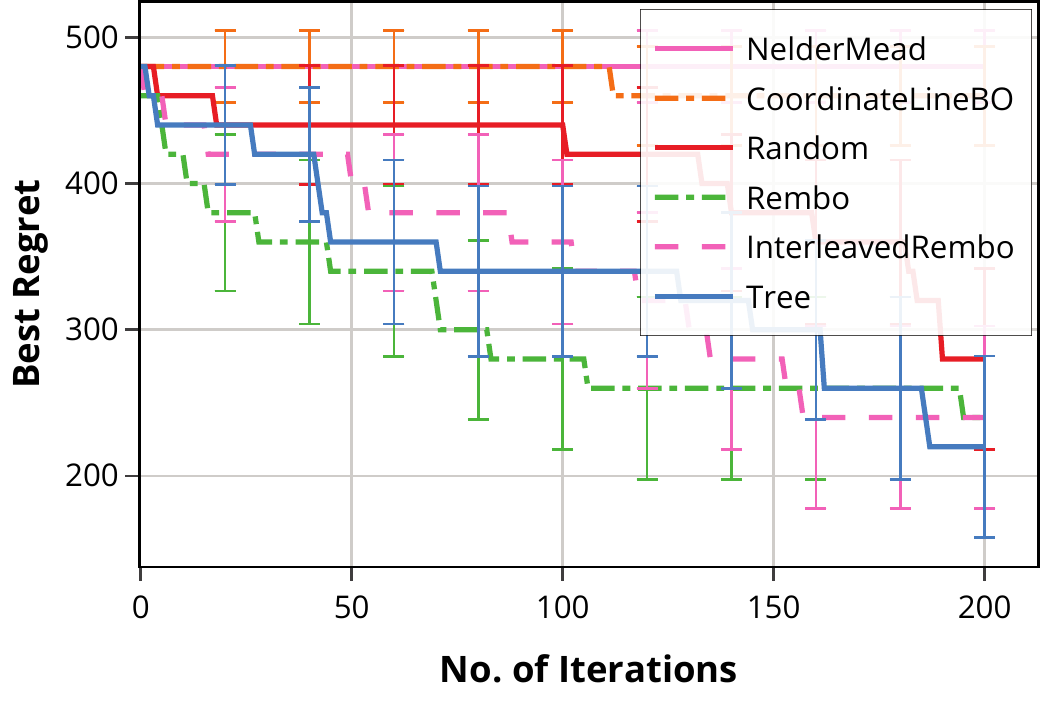} \label{fig:misc05inf-performance}}
    \caption{Comparison of various optimization algorithms for both synthetic functions and Lpsolve functions.}
\end{figure*}

\subsubsection{Discussion on Computation Time.} 
In general, it is difficult to compare the amount of computational resources by various algorithms, 
as it is very much affected by many factors such as implementation, hardware, underlying GP backends, etc. 
However, we provide a brief discussion of the general trends observed.
We generally found LineBO to be one of the faster approaches due to the use of 1D subroutines, 
though we also found its optimization perform to be limited in several cases. 
A fairly similar discussion applies to REMBO. 
On the other hand, the computational requirements of the {\em additive} methods are somewhat easier to compare in a fair manner, as we now discuss.

\subsubsection{Cost Efficiency.}
For the additive methods, we compute the \emph{Message Passing Cost} counting the number of individual acquisition function (\(\phi^G\)) evaluations; 
see (\refeq{eq:acquisition-psi}) in the appendix.
This metric is a proxy for the computational resources used in the optimization of the acquisition function. 
While it may not always correspond exactly to the total computation time, we expect that each message passing operation for \tree~is {\em at least} as fast as in \graphnooverlap~and \graphoverlap. 
This is because \tree~only works with functions containing only one or two variables, whereas the others may contain a larger number of variables.

\subsection{Experiments with Additive GP Functions} \label{sec:syn-add}
We first compare \tree~to other additive methods for functions drawn from a GP with additive structure. 
We focus our discussion on understanding the additive methods' scaling ability and performance. 
Afterwards, we compare \tree~to other methods using various non-GP functions. 
On all synthetic experiments, we add Gaussian $\mathcal{N}\left(0,{0.15}^2\right)$ noise to simulate noise that occurs in real-world applications.

Similar to \cite{rolland2018high}, we test our algorithm on synthetic data by sampling functions from GPs with several different additive dependency structures. 
We use an RBF kernel with corresponding dimensional lengthscale and scale parameters set to \(\sigma^\mathrm{opt}_i = 1\) and \(l^\mathrm{opt}_i = 0.2\) for all $i$.
We tested several dependency structures; three notable examples are illustrated in \refig{fig:synthetic}, and a full list is given in the appendix.

In \refig{fig:StarGraph-25-synd-performance}, it is unsurprising that \tree~outperforms the other additive methods for Star-25.
The dependency graph of Star-25 is indeed a tree, 
enabling our method to be effective in learning the dependency structure.
From \refig{fig:appendix-StarGraph-25-synd-cg} in the appendix, by plotting $\fscore$ over iterations,
we observe that it is efficient in learning the dependency structure. 
The dependency structure learned by \tree~is closest to the ground-truth throughout the experiment,
when compared with other additive models.
This efficiency is also reflected in \refig{fig:appendix-StarGraph-25-synd-cost}, 
where \tree~achieves the best performance as a function of the message passing cost.  
We additionally demonstrate in the appendix that the reduction in cost becomes significantly higher in the case of continuous domains, 
achieving better performance return on cost than other additive methods.

Next, we turn to the case that the underlying graph is not a tree. \refigs{fig:GridGraph-9-sync-performance}{fig:GridGraph-9-sync-cg} corresponds to the Grid-3$\times$3 structure, 
and we find that \graphoverlap~performs the best in terms of learning the dependency structure. 
This is because, for the grid graph model, only \graphoverlap's underlying structural assumptions are correct.
Both \tree~and \graphnooverlap~face difficulty learning the graph accurately, albeit worse for \graphnooverlap.
Interestingly, \tree~still remains competitive in terms of optimization performance despite poorer graph learning.
That is, when \tree~makes errant connections (or errant non-connections) in the dependency graph, 
the performance does not degrade significantly, and the algorithm can tweak other parameters 
(e.g., $\sigma_i$ and $l_i$) to minimize the effect of any errant connections.
From \refig{fig:GridGraph-9-sync-performance}, despite all additive algorithms being mutually competitive in terms of regret, 
both \graphnooverlap~and \graphoverlap~needed more acquisition function evaluations to achieve the same performance as \tree~(more than triple for \graphnooverlap);
see \refig{fig:appendix-GridGraph-9-sync-cost} in the appendix. 
In this instance, \graphnooverlap's pairwise disjoint assumption not only results in worse graph learning, but also worse cost efficiency.
Next, we compare \graphnooverlap~and \tree~using an Ancestry-132 dependency structure (132D).\footnote{See the appendix for more details on Ancestry-132.}
We found that \graphoverlap~was unable to complete such high-dimensional experiments in a reasonable time.
For \graphnooverlap~to work efficiently, we limited the maximum clique size, consider limits of both $5$ and $10$, represented by \graphnooverlapl{5} and \graphnooverlapl{10} respectively.
In \refigs{fig:AncestryGraph-132-sync_tree-performance}{fig:AncestryGraph-132-sync_tree-cost}, 
we find \tree~performing best in high-dimensions, and being the most cost efficient.

\paragraph{Scalability.} Here, we test \tree's scalability to higher dimensions up to 225D, 
focusing on studying how the total cumulative message passing cost incurred scales as dimension increases.
We used the same setup and parameters as \refsec{sec:syn-add}, 
across additive grid structures of varying sizes -- Grid-$i$$\times$$i$ for $i\in[2,15]$.
We again include \graphnooverlapl{5} and \graphnooverlapl{10} for this experiment.

From \refig{fig:scalability}, we can see that the amount of cost needed for \graphoverlap~and \graphnooverlap~quickly increases as the dimensionality increases.
In fact, we were unable to complete the experiment for larger grids in a reasonable amount of time.
Recalling that \graphoverlap~runs in time exponential in the size of the maximum clique of the triangulated dependency graph \cite{rolland2018high}, 
we note that even if that clique size does not grow large for the {\em true} graph, 
it may still tend to increase for the {\em estimated} graph.
Similarly, \graphnooverlap~may be slow due to the consideration of large cliques, unless the clique size is explicitly limited. 
Even after imposing the limits, we found that \tree~still incurs the lowest cost when compared with both \graphnooverlapl{5} and \graphnooverlapl{10}.
This is because, for tree structures, the message passing cost is quadratic in the number of discretization levels of a single dimension.

\subsection{Experiments with Non-GP Functions}
\subsubsection{Non-GP Synthetic Functions.}
Here we test our algorithm against commonly used BO synthetic function benchmarks \cite{oh2018bock,kirschner2019adaptive}, including Hartmann6 (6D) and Stybtang250 (250D). 
We also tested \tree~on benchmarks with invariant subspaces; 
following the setup in \cite{kirschner2019adaptive}, Hartmann6+14Aux (20D) was obtained by augmenting the synthetic functions with $14$ auxiliary dimensions.
In \refig{fig:Hartmann6Aug-hpoca-performance}, 
we see that the regret of \tree~reduces rapidly compared to other methods, 
with variants of LineBO catching up in later iterations. 
In \refig{fig:Stybtang-stybtang-performance}, we see that \tree~again manages to scale well in 
higher-dimensional synthetic functions.
From additional synthetic experiments (\refigs{fig:appendix-Camelback-hpoc-performance}{fig:appendix-Stybtang-stybtang-performance} in the appendix), 
\tree~is also competitive against LineBO variants across both lower and higher dimensional settings, 
even in cases with invariant subspaces.

\subsubsection{Linear Programming Solver.}
We consider tuning the parameters of lpsolve,
an open-source Mixed Integer Linear Programming (MILP) solver \cite{Berkelaar2004}.
The parameters within each algorithm typically have some relationship with each other; 
tweaking a parameter can potentially affect another.
We consider a similar configuration problem as defined by \cite{hutter2010automated,wang2013bayesian},
focusing on tuning lpsolve's 74 parameters - 59 binary, 10 ordinal and 5 categorical.
Our objective is to find the set of parameters of lpsolve that minimize the \emph{optimality gap} it can achieve with a time limit of five seconds
for the MIP encoding `misc05inf' found in the benchmark MIPLIB \cite{miplib2017}.

From \refig{fig:misc05inf-performance}, we observe that REMBO is competitive in performance for optimizing the linear programming solver,
as parameter optimization problems often have low effective dimensionality \cite{wang2013bayesian,hoos2014efficient}.
Despite being based on a very different notion of structure, \tree~attains better performance than REMBO in this example, with both clearly outperforming \random.
In the appendix, we provide two additional lpsolve examples in which \tree~outperforms REMBO.

\subsubsection{Additional Experiments.}
Additional experiments on the NAS-Bench-101 (NAS) dataset \cite{ying2019bench,klein2019tabular} and 
BO-based adversarial attacks (BA) \cite{Ru2020BayesOpt} can be found in the appendix.

\section{Conclusion}
For the problem of GP optimization with generalized additive models, 
we traded off expressivity for computational efficiency and ease of model learning by reducing the model complexity, constraining the dependency graph to tree structures.
Our method efficiently learns the additive tree structure using Gibbs Sampling and edge mutation,
suitable for resource-limited settings in line with the primary motivation of BO. 
Besides, we presented a zooming-based message passing approach that can benefit BO with generalized additive models in continuous domains, with or without tree structures.
We demonstrated that \tree~is competitive on both synthetic functions and real datasets, 
and that the computation can be significantly reduced compared to more complex graph structures, without sacrificing the optimization performance.

\section*{Acknowledgments}
This work was supported by both the Singapore National Research Foundation (NRF) under grant number R-252-000-A74-281 and the AWS Cloud Credits for Research program.

\fontsize{9.0pt}{10.0pt}
\selectfont

\includepdf[pages={-}]{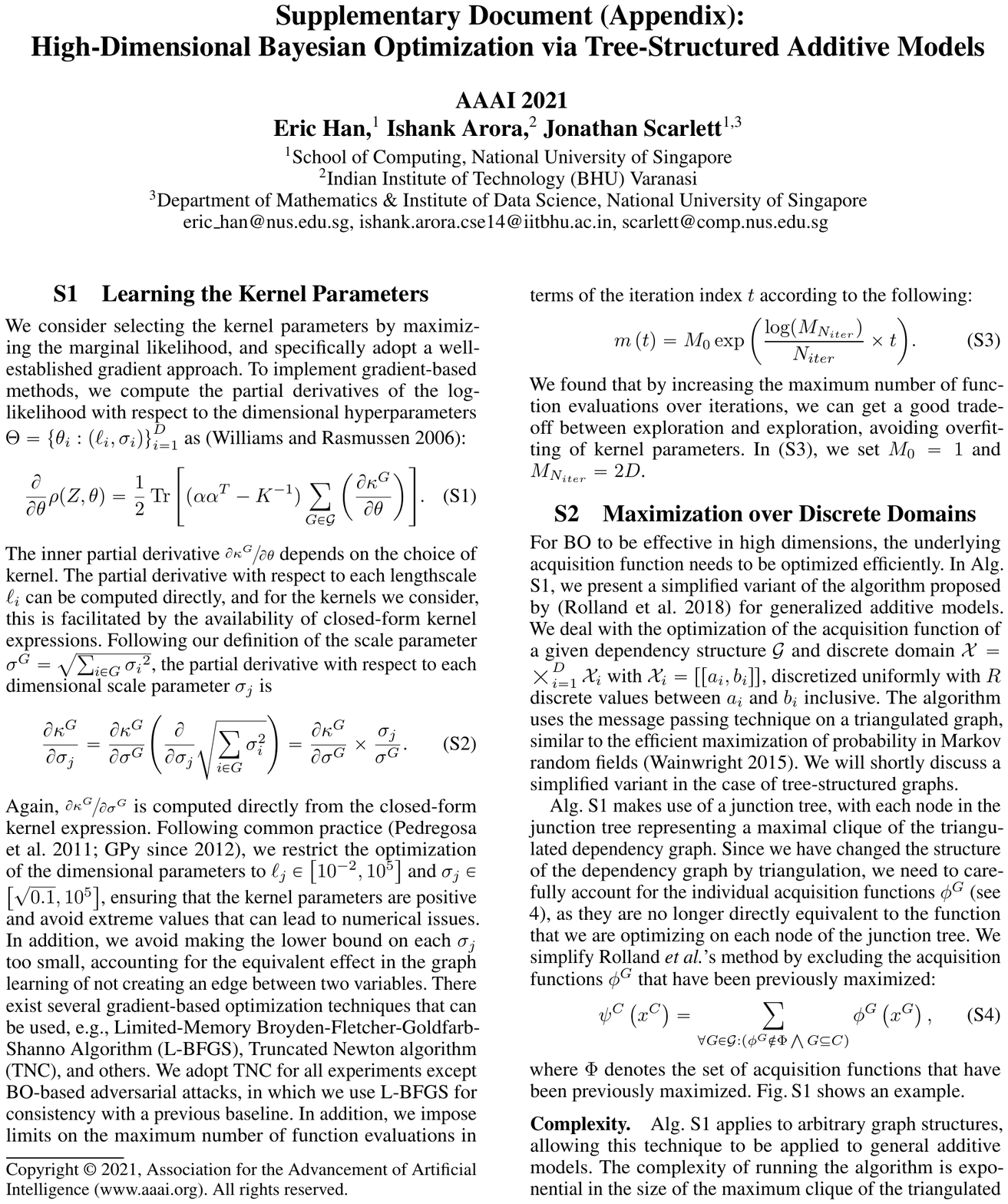}
\end{document}